\documentclass[lettersize,journal]{IEEEtran}
\usepackage{amsmath,amsfonts}
\usepackage{algorithmic}
\usepackage{algorithm}
\usepackage{array}
\usepackage[caption=false,font=normalsize,labelfont=sf,textfont=sf]{subfig}
\usepackage{textcomp}
\usepackage{stfloats}
\usepackage{url}
\usepackage{verbatim}
\usepackage{graphicx}
\usepackage{cite}
\usepackage{wrapfig}

\usepackage{amsmath}
\usepackage{multirow}
\usepackage[nottoc]{tocbibind}
\usepackage{float}
\usepackage{soul}
\usepackage[justification=centering]{caption}
\usepackage[english]{babel}
\usepackage{colortbl}
\usepackage{amsfonts}
\usepackage{graphicx}
\usepackage{lipsum}
\usepackage{multicol}
\usepackage{cite}
\usepackage{booktabs}
\usepackage[referable]{threeparttablex}
\usepackage{algorithm}
\usepackage{algorithmic}
\usepackage[dvipsnames]{xcolor}
\definecolor{objcolor}{rgb}{204,204,255}
\usepackage{subfiles}
\usepackage{nomencl}
\sethlcolor{yellow}
\usepackage{xcolor}
\makenomenclature

\newcolumntype{L}{>{\centering\arraybackslash}m{3cm}}
\newcolumntype{S}{>{\centering\arraybackslash}m{1.5cm}}

\begin{document}

\title{SalienDet: A Saliency-based Feature \\ Enhancement Algorithm for Object \\Detection for Autonomous Driving}


\author{Ning Ding, Ce Zhang \textit{IEEE Member} and Azim Eskandarian \textit{IEEE Senior Member}}

\markboth{Journal of \LaTeX\ Class Files,~Vol.~14, No.~8, August~2021}%
{Shell \MakeLowercase{\textit{et al.}}: A Sample Article Using IEEEtran.cls for IEEE Journals}


\maketitle

\begin{abstract}

Object detection (OD) is crucial to autonomous driving. On the other hand, unknown objects, which have not been seen in training sample set, are one of the reasons that hinder autonomous vehicles from driving beyond the operational domain. To addresss this issue, we propose a saliency-based OD algorithm (SalienDet) to detect unknown objects. Our SalienDet utilizes a saliency-based algorithm to enhance image features for object proposal generation. Moreover, we design a dataset relabeling approach to differentiate the unknown objects from all objects in training sample set to achieve Open-World Detection. To validate the performance of SalienDet, we evaluate SalienDet on KITTI, nuScenes, and BDD datasets, and the result indicates that it outperforms existing algorithms for unknown object detection. Notably, SalienDet can be easily adapted for incremental learning in open-world detection tasks. The project page is \url{https://github.com/dingmike001/SalienDet-Open-Detection.git}.  
\end{abstract}

\begin{IEEEkeywords}
Saliency Map, Open-World Object Detection, Cross-Data Evaluation, Autonomous Driving, Machine Learning
\end{IEEEkeywords}


\nomenclature{$S$}{Saliency Map}
\nomenclature{$I$}{Image Tensor}
\nomenclature{$I_{S}$}{Image Tensor Merged with Saliency Map}
\nomenclature{$P_{o}$}{Potential Object Proposals from SalienDet}
\nomenclature{$P_{g}$}{Ground Truth Proposals}
\nomenclature{$P_{e}$}{Proposals from Prior Knowledge}

\printnomenclature
\section{Introduction}

Object detection (OD) is crucial for autonomous vehicle systems, serving as vehicles' eyes to aid the control system in making informed decisions, such as acceleration, lane changes, etc \cite{eskandarian2019research, 10.1115/IMECE2021-69975, Mehr2022XCARAE}. 
The current autonomous vehicle systems employ various crucial object detection (OD) methods, among which the notable families are R-CNN \cite{girshick2014rich, girshick2015fast, ren2015faster, he2017mask, cai2018cascade, pang2019libra} and YOLO \cite{redmon2016you, redmon2017yolo9000, redmon2018yolov3, bochkovskiy2020yolov4}.
However, it is crucial to note that current autonomous vehicle OD applications are predominantly reliant on Close-Set Object Detection, which means they are limited to detecting only the objects learned during the training process \cite{merrill2020modified}. 
Such limitation becomes a significant issue in the development of autonomous vehicles due to the unpredictable nature of the real-world driving environment.
An example is when an unlabeled construction vehicle appears in front of the ego vehicle. 
Since such an object never appears in the training set, the ego vehicle's perception system fails to detect it, which leads to a failure in the control system to issue an order to avoid a collision, posing a serious safety hazard.\begin{figure}[h]
    \centering
    \captionsetup{justification=justified}
    \includegraphics[width=8.5cm]{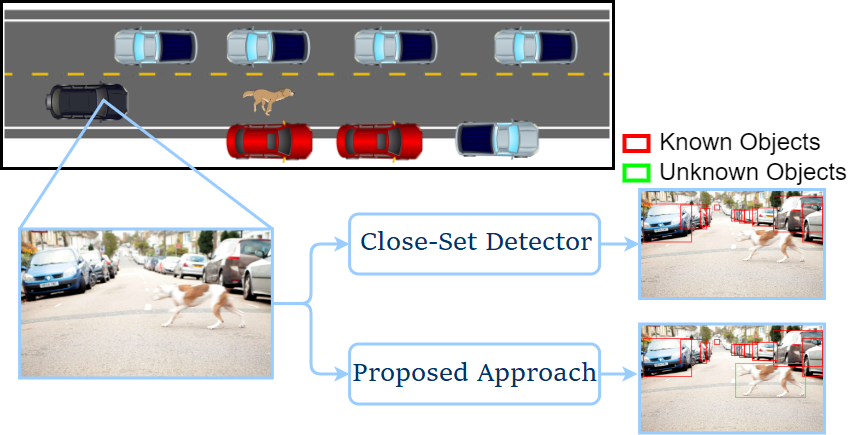}
    \caption{Close-Set detector only detects the classes which have been learned during the training process. The issue if the detector can catch the unknown classes which have never been learned before has become a new challenge for OD (Open-Set or Open-World). In this paper, we introduce the additional saliency map to the OD problem. We prove it could be a solution to help the detector to catch unknown objects.}
    \label{fig:Framework}
\end{figure}
Thus, enabling OD applications to detect unlearned objects would significantly improve autonomous vehicles safety.

The finding of unlearned objects is initially defined as an Open-Set Recognition problem\cite{scheirer2012toward}, which is later expanded to Open-World Recognition\cite{bendale2015towards} by incorporating the labeling of new classes and incremental learning. 
\cite{scheirer2012toward} and \cite{bendale2015towards} limit their studies in single object classification.
\cite{miller2018dropout} uses dropout sampling in OD under Open-Set condition and opened the door to Open-Set Detection.
Joseph et al. \cite{joseph2021towards} takes this a step further by extending previous work to new problem, namely Open-World Object Detection, which involves detecting new object classes and updating the detector with newly detected objects. 
However, Joseph and other Open-World related research\cite{wu2022uc,zohar2022prob,kim2022learning} tend to focus on generic datasets like MS-COCO \cite{lin2014microsoft} and PASCAL-VOC\cite{everingham2010pascal}, which include images with numerous object classes, relatively large object scales, and are taken in favorable conditions. 

Compared with generic datasets, autonomous driving datasets present unique challenges, including fewer object classes (traffic-only classes), more small-size objects, and diverse weather conditions. 
More specifically, a limited number of object classes in the dataset can significantly impact the ability of the detector to learn the general characteristics of potential objects, ultimately affecting its performance in Open-World Object Detection. 
It is also challenging for the detector to detect objects accurately when presented with images containing objects that are too small to be perceived by the human eye or low-quality images captured under poor weather conditions.

Based on the aforementioned issues, we propose a novel algorithm, named as SalienDet, for unknown object perception in an autonomous driving environment.
SalienDet integrates saliency map to enhance the image features so that both known and unknown objects are highlighted.
The contributions of our study are shown below:

\begin{itemize}
\item We investigate and prove the potential of spectral residual-based saliency map \cite{hou2007saliency} on object features enhancement.
\item We integrate the saliency map with a customized training paradigm to enable the detector to capture unknown objects (saliency Open-Set).
\item Our research expands the saliency Open-Set studies to Open-World detection (saliency Open-World).
We utilize the SalieDet to reconstruct the data labels.
Then, a regular OD algorithm is trained based on the newly constructed dataset, which can learn both known and unknown objects in an incremental manner.
\item SalienDet is evaluated on three large-scale autonomous driving datasets to prove the model's robustness and correctness. To the best of our knowledge, this is the first time such a comprehensive evaluation has been carried out.
\end{itemize}

\section{Related Works}
\subsection{Close-Set Object Detection}
Famous neural network models for Close-Set Object Detection can be categorized into convolutional neural network (CNN) detectors and transformer-based detectors. 
\subsubsection{CNN Approach with Region Proposals}
R-CNN\cite{girshick2014rich}, Fast R-CNN\cite{girshick2015fast} and Faster R-CNN\cite{ren2015faster} are all region-based convolutional neural network models. 
R-CNN is considered the pioneer of this method. 
It uses selective search\cite{uijlings2013selective} to produce a set of possible object locations (region proposals). 
Then, R-CNN extracts the proposed regions' features with CNNs.
Finally, the classification and bounding box refinement of the proposed regions are accomplished using Support Vector Machines (SVM). 
Although accurate detection performance, the selective search process is time-consuming and cannot achieve real-time detection.
The successors of R-CNN, such as fast R-CNN, and faster R-CNN are aimed to solve the computation time issue.
Fast R-CNN extracts features from the entire image instead of each region proposal, and replaces SVM in R-CNN with a series of fully connected layers,  resulting in a faster computation time. 
Inspired by Fast R-CNN, faster R-CNN further improves the speed and accuracy by discarding selective search and using an additional convolutional network, named as Region Proposal Network (RPN), to generate region proposals. 
This RPN shares the convolutional feature with the OD network, leading to a further reduction in processing time.
 In addition to above three approaches, the R-CNN family also includes Mask R-CNN\cite{he2017mask}, Cascade R-CNN\cite{cai2018cascade}, Libra R-CNN\cite{pang2019libra}, and other related models that share the same fundamental principles and use CNN to extract features and perform object detection on region proposals. 
\subsubsection{CNN Approach without Region Proposals}
YOLO\cite{redmon2016you} redefines OD as a regression problem by discarding the idea of region proposals. 
Instead, it uses a single neural network to predict both the class probability and the bounding boxes of multiple objects, making it faster and simpler than region-based convolutional neural network models. 
Despite its strong performance, YOLO has limitations in detecting small objects, prompting the development of newer versions to address this challenge. 
YOLOv2\cite{redmon2017yolo9000} introduces anchor boxes to improve the detection of objects of varying scales, YOLOv3\cite{redmon2018yolov3} uses a Feature Pyramid Network (FPN) to improve the handling of small objects, and YOLOv4\cite{bochkovskiy2020yolov4} leverages a more powerful backbone architecture called CSPDarknet53 to further enhance the model's ability to detect small objects while maintaining high performance.

\subsubsection{Transformer-based Approach}
As the effectiveness of transformer-based approaches has been demonstrated in handling variable tasks \cite{9564709}, current studies are incorporating transformers into OD tasks.
DETR\cite{carion2020end} utilizes a transformer-based architecture to encode the complete feature map extracted from the input image and generates a set of object queries, which are used to predict the presence and location of objects. 
The object queries are then matched with the encoded information using an attention mechanism, which enables the model to selectively attend to the most relevant features for each object query.
Finally, the attended features are used to predict the bounding box and class probabilities for each object in the image by a set of feedforward layers.
Although being a much simpler and more flexible architecture, DETR is challenging to train and has difficulty in detecting smaller objects. 
To overcome these challenges, Deformable-DETR\cite{zhu2021deformable} is introduced as an enhanced version of DETR, incorporating deformable attention modules to modify its attention mechanism. By utilizing a limited number of keys for each query, Deformable-DETR prevents the model from becoming overly complex and facilitates easier convergence. In addition, the use of multi-scale feature maps allows Deformable-DETR to perform well in detecting smaller objects.
\subsection{Open-World Object Detection}
Joseph\cite{joseph2021towards} formalizes Open-World Object Detection, which connects Open-Set Detection with incremental learning. 
He proposes a solution to this problem called ORE (\underline{O}pen Wo\underline{r}ld Object D\underline{e}tector). 
The ORE model determines whether an object is unknown by calculating the energy of its potential object feature vector. 

After Joseph, several other researchers begin to work on the problem of Open-World Object Detection.
One such approach used by some researchers is the use of unsupervised classifiers to handle unknown classes. Xiaowei Zhao\cite{zhao2022revisiting} proposes a Class-specific Expelling Classifier that expels ambiguous instances from the predicted known class and reassigns their class predictions to other classes. 
Another approach is to use end-to-end neural network to solve the problem. 
Kim\cite{kim2022learning}'s Object Localization Network (OLN) is based on Faster R-CNN.
He replaces the classifier head in the RPN stage and RoI stage with localization quality estimators, and shows that this method can effectively propose new object proposals. 
Singh's research \cite{singh2021order} builds upon Zhong's work \cite{zhong2021openmix} by proposing a method called Feature-Mix.
This involves mixing the features of prior knowledge of known classes and query objects, which has the potential to reduce the interference from known classes to query objects and consequently enhance the classifier's ability to identify unknown objects. 
Gupta\cite{gupta2022ow} makes modifications to the Deformable-DETR pipeline by selecting high objectiveness score queries that are not matched to known classes as matches for unknown classes. 
This approach, named as Attention-driven Pseudo-labeling, allows the subsequent classifier to classify objects that are significantly different from the background into a new class.
Ma\cite{ma2023cat} develops the method that integrates Gupta's Attention-driven Pseudo-labeling with selective search to improve the selection of queries for unknown classes.
While the approaches discussed above have made significant contributions to the field of Open-World Object Detection, they do not directly address the problem of emphasizing the unknown objects in the overall dataset. 
Highlighting these objects can theoretically enable the classifier to more easily distinguish and identify unknown classes, and this is the primary focus of our work.

\subsection{Saliency Map}
A saliency map is a type of heat map that highlights the most visually significant objects or regions in a given visual scene. 

Due to the immense power of neural networks, many researchers have utilized them to generate saliency maps for solving computer vision problems. 
Hong\cite{hong2015online} achieves high performance in object tracking by using the last convolutional layer of a CNN to compute the saliency map specific to the target object. 
Zhao\cite{zhao2011learning} employs a least squares technique to learn the weights between the feature maps generated by a neural network. 
These learned weights can be used to enhance the quality of the final saliency map by improving the original feature maps.
Simonyan\cite{simonyan2013deep} utilizes a classification network to extract a saliency map, which is then incorporated into a weakly supervised object segmentation approach. 
Rui Zhao's method\cite{zhao2015saliency} is a multi-context deep learning framework that incorporates both global and local contexts to produce high-quality saliency maps. 

Despite the significant contributions made by these works, they rely heavily on large datasets and prior knowledge, which may improve the performance of detecting learned classes but provide less assistance in detecting novel classes.

Conversely, several earlier works do not rely on any prior knowledge of objects and instead utilize internal characteristics such as color, contrast, texture, and edge detection to compute the intensity area in an image.
This approach is more cost-effective and enables researchers to extract more information from the images.
Montabone\cite{montabone2010human} utilizes a center-surround saliency mechanism to calculate intensity maps, which enables the direct extraction of visual features from the image. Walther\cite{walther2002attentional} generates an intensity conspicuity map by identifying areas in an image that contribute most to saliency based on colors, intensities, and orientations. 
Hou's spectral residual approach\cite{hou2007saliency} extracts objects from their backgrounds by analyzing the spatial domain in each image. This method helps to identify and isolate specific objects within an image.

The lack of a need for prior knowledge in early saliency map generating methods offers an advantage that inspires the integration of such methods into neural network detectors. 
By doing so, it enhances the feature representation of objects, making detection easier. Moreover, since prior knowledge is not required for generating the saliency map, it enhances the feature representation of all possible objects, even those not labeled in the training set. As a result, the detector becomes capable of discovering more object classes. This is particularly useful for the Open-World Object Detection problem, as early methods can address the issue of grouping objects outside of prior knowledge as background.

We opt to use the spectral residual approach proposed by Hou to generate the additional saliency map in this study. 
This method explores the characteristics of backgrounds and is capable of capturing more objects in the resulting saliency map with a relatively efficient computation load.

\section{METHOD}

The objective of OD is to detect target objects with a given image frame. 
In Close-Set Detection, objects that are not present in the training sample set are treated as background and their information is typically suppressed in the feature map during training.
However, in Open-Set and Open-World Detection, the challenge is to detect both known and unknown object classes, as the network needs to capture features of novel classes that are not present in the training sample set, while still distinguishing them from the background.
Based on such requirements, if we could incorporate additional highlighted information of other objects in the original picture to guide the detector's attention, it would be helpful for the detector to pay attention to highlighted objects in additional salicncy map as well as labeled objects in the original images. 
To achieve this goal, we utilize spectral residual-based saliency maps as an additional source of information and propose the approach SalienDet. 
Compared with other methods, the spectral residual-based saliency approach offers several advantages: a) it is computationally efficient, making it suitable for real-time applications when compared to other methods \cite{li2013saliency,duan2011visual,feng2011salient}; b) it focuses on the spectral characteristics of the image, which enables it to maintain robustness against changes in lighting conditions, colors, and other image variations. This characteristic is particularly beneficial for autonomous vehicle datasets that are often captured under poor weather conditions; c) the effectiveness and practicality of this approach has been demonstrated in various applications \cite{liu2016saliency, li2013object}, where it has been utilized as a preprocessing step for computer vision tasks.

As shown in Fig \ref{fig:Framework}, SalienDet includes the process merging saliency map with images and a detector. 
In our study, we first apply SalienDet in a Close-Set environment to examine the impact of incorporating a saliency map on OD. 
After observing a positive effect on OD, we then assess its effectiveness in an Open-Set problem by testing whether it can detect novel object classes.
Finally, we integrate SalienDet into the dataset modification process to address the Open-World problem. Overall, our study explore the applicability of SalienDet to various OD challenges.
\begin{figure}[h]
    \centering
    \captionsetup{justification=justified}
    \includegraphics[width=8.5cm]{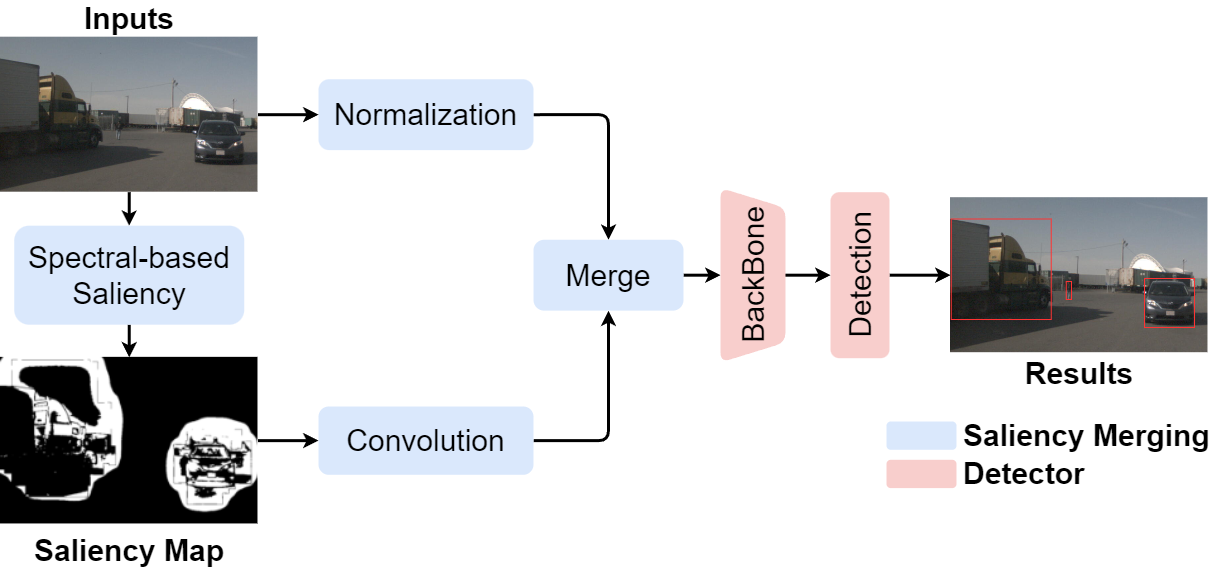}
    \caption{Overview of SalienDet. The SalienDet model uses a new input for the backbone that combines the original image and its saliency map to detect objects based on information from both the original image and the additional saliency map.}
    \label{fig:Framework}
\end{figure}

\subsection{Merge Saliency Map with Images}

A saliency map $S$ can be generated either for the entire image or only for specific regions that are deemed important.

Following the generation of the saliency map, we employ a two-layer convolutional approach to augment the channels in $S$ from one to three. 
We subsequently merge the initial image tensor $I$ with the newly-formed three-channel saliency map to construct a new image tensor $I_{S}\in \mathbb{R}^{H\times W\times 3}$, which replaces the original image as the input to the detector's backbone. Such action can be described as:

\begin{equation}
I_{S} = I + b_{3}+ w_{3}\star (b_{1}+w_{1}S)
\end{equation}
Where, $b_{1}$ and $b_{3}$ are learnable biases for 1-channel tensor and 3-channel tensor.  $w_{1}$ and $w_{3}$ are learnable weights for 1-channel tensor and 3-channel tensor.  $\star$ represents a 2-dimensional cross-correlation operator.

\subsection{Detect Objects for Close-Set} 
Considering that labeled information will guide the detector to ignore the unlabeled objects even though they are highlighted by other methods, we convert the original whole images into saliency maps during training and inference. 
After generating the saliency map, it is merged with the original image according to the instructions in Section A to produce a new image tensor $I_S$. This modified image tensor is then used as the input for the detector during object detection.
In our study we choose Faster R-CNN as the detector. 

\subsection{Detect Objects for Open-Set}
The purpose of this step is to output potential object proposals $P_{o}$ based on the new image tensor $I_{S}$. 
We use Faster R-CNN as the detector to train, using the new image tensor $I_{S}$ as inputs. 
During the training process, the additional saliency map is only produced on the area of annotated known objects. 
The new image tensor $I_{S}$ only contains the additional highlight information of known objects.

As the generated saliency map may highlight objects not present in the dataset, it is not possible to assess the model's performance using the ground truth. 
The objective of this study is to demonstrate the effectiveness of saliency maps in addressing new class OD challenge, rather than maximizing the detection of new objects in the evaluation results. 
Our focus is primarily on the annotated objects present in the dataset. 
Therefore, we follow a similar approach used in other Open-World approach\cite{joseph2021towards,singh2021order}. we utilize a small sample set from the training data, which has been held out and referred to as the \textit{full-instances proposal sample set}, as prior knowledge for generating additional saliency maps.
Prior to performing inference,  we train a simple neural network on this sample set containing labeled known and unknown objects. 
The purpose is to obtain possible proposals $P_{e}$ for the annotated objects in the dataset by this simple neural network for every image in the test set. 
As the sample set used for prior knowledge is very limited, the resulting proposals $P_{e}$ only provide approximate bounding-box areas that may contain the relevant objects, instead of precise bounding-box inference results.

During inference, the saliency map is generated based on proposals $P_{e}$, with the aim of having the new image tensor $I_{S}$ highlight all known and unknown objects in the test set, aided by the additional saliency map.
Since no classification is performed at this step, all classes in the dataset are considered as a single group and treated as one class. 
The output of this step is the object proposals, represented by $P_o$.

This approach can be evaluated separately using a set of new images to determine its effectiveness in detecting object proposals, as demonstrated in Part B of Section VI. 
Alternatively, it can be integrated as a part of solution to Open-World Object Detection, producing potential object proposal images in the training set that can be used for dataset relabelling, as discussed in Part D of this Section.

\subsection{Modify Dataset for Open-World}
\begin{figure*}[h]
    \centering
    \captionsetup{justification=justified}
    \includegraphics[width=17.5cm]{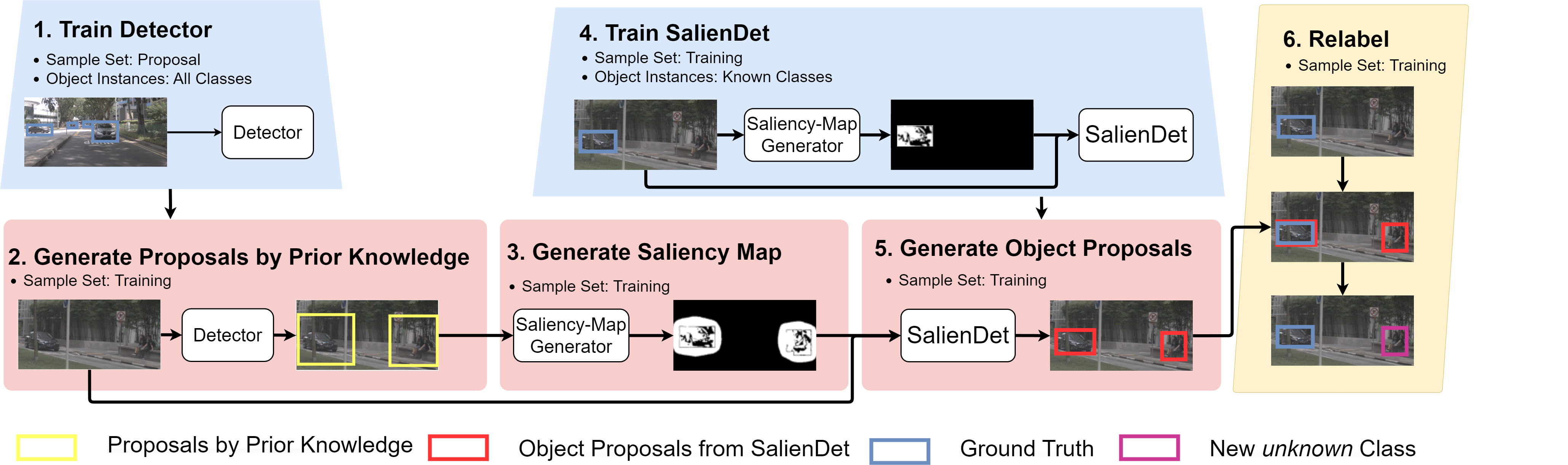}
    \caption{Process to modify the dataset for Open-World Object Detection problem. In this figure, trapezoid shapes indicate the model training phase, rectangle shapes represent the proposal generation phase, and parallelogram shape indicates the relabel phase. 1) A simple neural network is trained on proposal sample set. 2) The trained neural network model with prior knowledge is used to generate extra proposals. 3) The saliency map is generated based on extra proposals. 4) Only the labeled objects’ information is used to generate an additional saliency map, which is then used with original image to train the SalienDet. 5) The trained SalienDet generates objects proposals by the saliency map from step 3 and original image. 6) The object proposals from step 5 and ground truth are used to label unknown class.}
    \label{fig:train_and_inference_big}
\end{figure*}

Fig \ref{fig:train_and_inference_big} shows the process of utilizing SalienDet to modify the dataset for addressing the Open-World Object Detection problem.
It is worth noting that the primary objective of this method is to reconstruct the dataset and relabel objects' classes.
Once relabeling process is complete, any object detector can be used to achieve Open-World Object Detection.

For each incremental learning task of Open-World Object Detection, we begin by training the SalienDet using the images from the training sample set. 
During the training process, we only generate the saliency map for the image region containing labeled objects. 

Once the training is complete, we freeze the parameters of SalienDet. 
The trained SalienDet then utilizes the original training set images along with the saliency map generated from extra proposals $P_{e}$ (obtained from prior knowledge) to predict potential object proposals $P_{o}$.
Finally, we relabel the training set using the proposed $P_o$.

\begin{algorithm}
\caption{Label \textit{unknown} objects}\label{euclid}
\textbf{Input}: Object proposals $P_{o}$, known objects' ground truth proposals $P_{g}$, proposal threshold $\alpha$

\begin{algorithmic}[1]
\FOR{$\text{each object proposal in }P_{o}$}
\STATE {\textit{k} = False}
\FOR{\text{each ground truth proposal in }$P_{g}$}
\STATE {Compute \textbf{IoU} between each Object proposal and each ground truth proposal}
\IF{$IoU>\alpha$} 
\STATE{$\textit{k}=True$}
\ENDIF
\ENDFOR
\IF{$\textit{k}==False$}
\STATE {label object proposal as \textbf{unknown}}
\ENDIF
\ENDFOR
\end{algorithmic}
\end{algorithm}
Algorithm 1 shows the relabeling process. 
Intersection over Union (IoU) metric, which measures the ratio of overlap between two regions, is utilized in this step. 
We compare the object proposals $P_{o}$ with ground truth proposals. 
If the IoU between a proposal in $P_{o}$ and any proposal in ground truth is greater than the threshold $\alpha$, the object proposal is labeled as a known class. 
A for-loop is used to check each proposal in $P_{o}$, and any object proposals that do not meet the criteria for known class are labeled as unknown.
The proposal threshold $\alpha$ serves as a tuning parameter to select the most suitable proposals for unknown class. 
If the alpha $\alpha$ is lower, it becomes harder to classify object proposals as unknown because they have a lower overlap with the ground truth for known classes.
In this work, we have empirically set $\alpha=0.3$.

\section{Dataset and Reconstruction}
This study undergoes training and assessment on three renowned open-source datasets (KITTI, nuScenes, and BDD100K) independently. 
We utilize the datasets to address different detection problems Close-Set, Open-Set and Open-World), and we resample the dataset accordingly for each problem.

For performing Close-Set Object Detection, we employ the original labeled samples in the dataset. 

To conduct Open-Set Object Detection, we treat some classes as known class, the remaining classes in the dataset as unknown class. 
As present in Part C and Part D in Section III, we held out a small full-instances proposal sample set as prior knowledge for the additional saliency map generating.
We select only 500 samples with all classes labeled from the training sample set.

The Open-World Object Detection sample organization is similar to the Open-Set but with incremental learning capabilities.
We follow the data splitting method described by Joseph\cite{joseph2021towards} and divide the dataset into multiple tasks, denoted by $T  =\left \{  T_{1},...,T_{t},...\right \}$. 
At a particular time $t$, the model is presented with new classes, and all the classes in $T_{\tau}$ ($\tau<t$) are treated as known classes, while all the classes in $T_{\tau}$ ($\tau>t$) are regarded as unknown classes.
During each task, only the newly introduced classes are labeled during training, while all known classes are labeled, and unknown classes are labeled as unknown during validation.
We also keep 500 samples' proposal sample set aside as prior knowledge, similar to how it is done in Open-Set Object Detection. 
To prevent the detection model from forgetting previous known classes, images containing at least 50 instances per class \cite{joseph2021towards} are chosen as the exemplar replay set in each task to enhance the memory of the detector. 
This is based on the understanding that the detection model tends to forget previous known classes if it is only trained with newly introduced classes, which is known as catastrophic forgetting\cite{french1999catastrophic}.

It is crucial to note that this modification is carried out on the original labeled training sample set and labeled validation sample set separately.
Additionally, to guarantee a fair comparison, the performance of other object detection algorithms are also evaluated on the same sample set configuration as the proposed SalienDet model in Open-Set and Open-World Object Dection.

\subsection{KITTI}
The KITTI \cite{Geiger2012CVPR} Vision Benchmark Suite is a widely-used dataset in autonomous vehicle research \cite{zhao2019object,grigorescu2020survey}, with a section specifically dedicated to object detection. 

Table I displays the official KITTI dataset, which contains training sample set and test sample set. The dataset consists of 8 object classes, including car, van, truck, pedestrian, person sitting, cyclist, tram, and misc (e.g. trailers). 
Because the official test sample set does not have ground truth, we split the official training sample set into a new training sample set and a new validation set. 
These two new sets are created by randomly selecting images from the official training sample set. 
The samples and instances for these new sets can also be found in Table \ref{tab:official_kitti}.
\setcounter{table}{0}
\begin{table}[H]
\centering
\caption{Official and Split Samples \\ and Instances Statistics in KITTI Dataset}
\begin{tabular}{c|c|c|c}
\toprule[1.3pt]
\midrule[0.3pt]
\multicolumn{4}{c}{\textbf{Samples}}                             \\\midrule

\multicolumn{2}{c|}{\textbf{Official}} & \multicolumn{2}{c}{\textbf{Split}}                            \\\midrule

\textbf{Training}    & \textbf{Testing} & \textbf{Training}    & \textbf{Validation}    \\\midrule
7,481               & 7,518   & 5,984 & 1,497                    \\\midrule[1.1pt] 
\multicolumn{4}{c}{\textbf{Instances}}                           \\\midrule
  & {\textbf{Official}} & \multicolumn{2}{c}{\textbf{Split}}                             \\\midrule

\textbf{Classes}     & \textbf{Training}  & \textbf{Training}   & \textbf{Validation}    \\\midrule
car              & 28,741  & 23,174 & 5,567                \\\midrule
van           & 2,912   & 2,344 & 568           \\\midrule
cyclist                  & 1,627 & 1,291 & 336              \\\midrule
pedestrian                 & 4,486    & 3,630 & 856          \\\midrule
truck         & 1,094 & 900 & 194              \\\midrule
tram                 & 511 & 410 & 101              \\\midrule
misc                & 972 & 802 & 170             \\\midrule
person sitting                & 222  & 172 & 50             \\\bottomrule[1.1pt]
\end{tabular}
\label{tab:official_kitti}
\end{table}

The results of our resampling approach for three different detection problems are presented in Table \ref{tab:kitti_resample}.

For performing Close-Set Object Detection, we employed the split sampling setup depicted in Table \ref{tab:official_kitti}. 
To conduct Open-Set Object Detection, we treated the \textbf{car} and \textbf{truck} classes as known objects, whereas the other 6 classes were treated as unknown. In the Open-World Object Detection approach, the dataset is resampled for three tasks, and for each task, a set of newly introduced classes is defined. In Task 1, the newly introduced classes are \textbf{car} and \textbf{truck}. In Task 2, the newly introduced classes are \textbf{tram}, \textbf{misc}, and \textbf{cyclist}. Finally, in Task 3, the newly introduced classes are \textbf{pedestrian} and \textbf{van}.

\subsection{nuScenes Dataset}
The nuScenes\cite{nuscenes2019} dataset is a relatively new and more challenging dataset compared to KITTI. 
The object detection section of the dataset, called nuImages, includes training, validation, and testing sample sets with 15 object classes. The object classes includes: pedestrian, barrier, debris, pushable-pullable objects, traffic cone, bicycle rack, bicycle, bus, car, construction vehicle, emergency vehicle, motor, trailer, and truck.

We merge some similar classes into a single object class. 
Specifically, we combine \textbf{bicycle} and \textbf{bicycle rack} into the class \textbf{bike}, while \textbf{debris} and \textbf{pushable-pullable objects} were merged into the class \textbf{road objects}. \textbf{Emergency vehicle} is included into \textbf{car}. 
We also exclude the class \textbf{animal}, which has a very small number of instances and is rarely encountered in driving conditions.
The final samples and instances of each merged class are in Table \ref{tab:nu_official}.

We utilize samples in training sample set and validation sample set, scince the test sample set does not contain any ground truth. Table \ref{tab:nu_resample} shows the resampling result of the dataset.

For Close-Set Object Detection, we use the same sampling setup as the Table \ref{tab:nu_official}.
For Open-Set, \textbf{car} and \textbf{pedestrian} classes are treated as known objects, while the remaining classes are considered unknown.
For the Open-World, the introduced classes for Task 1 are \textbf{car} and \textbf{bus}. 
In Task 2, the newly introduced classes are \textbf{motor}, \textbf{bike}, \textbf{barrier}, \textbf{traffic cone} and \textbf{road objects}.
In Task 3, the newly introduced classes are \textbf{trailer}, \textbf{truck}, \textbf{construction vehicle} and \textbf{pedestrian}.
\setcounter{table}{1}
\begin{table}[H]
\centering
\caption{Samples and Instances Statistics \\ in nuScenes Dataset}
\begin{tabular}{c|c|c}
\toprule[1.3pt]
\midrule[0.3pt]
\multicolumn{3}{c}{\textbf{Samples}}                             \\\midrule
\textbf{Training}    & \textbf{Validation} & \textbf{Testing}    \\\midrule
67,000               & 16,000              & 10,000              \\\midrule[1.1pt] 
\multicolumn{3}{c}{\textbf{Instances}}                           \\\midrule
\textbf{Classes}     & \textbf{Training}   & \textbf{Validation} \\\midrule
barrier              & 70,112              & 18,433              \\\midrule
pedestrian           & 135,870             & 32,710              \\\midrule
car                  & 202,947             & 47,322              \\\midrule
bike                 & 16,169              & 3,955              \\\midrule
traffic cone         & 69,016              & 18,587              \\\midrule
bus                  & 6,741               & 1,885               \\\midrule
truck                & 29,456              & 6,858               \\\midrule
motor                & 13,682              & 3,097               \\\midrule
road objects         & 5,491               & 1,355               \\\midrule
construction vehicle & 4,768               & 1,303               \\\midrule
trailer              & 3,285               & 486                 \\\bottomrule[1.1pt]
\end{tabular}
\label{tab:nu_official}
\end{table}

\subsection{BDD100K}
\setcounter{table}{3}
\begin{table*}[hpb]
\centering
\caption{KITTI Dataset Resampling}
\begin{tabular}{S|S|L|L|L|L}
\toprule[1.3pt]
\midrule[0.3pt]
\multicolumn{6}{c}{\textbf{Samples}}                                                                                                                                                                                                                                                                                                                                      \\ \toprule[1.3pt]
                           & \textbf{Close-Set} & \textbf{Open-Set}                                                                          & \multicolumn{3}{c}{\textbf{Open-World}}                                                                                                                                                                                    \\\midrule
                           & \textit{Task}      & \textit{Task}                                                                              & \textit{Task 1}                                                                            & \textit{Task 2}                                                     & \textit{Task 3}                                         \\\midrule
\textbf{Training}          & 5,984             & 5,484                                                                                     & 5,400                                                                                     & 5,008                                                                  & 5,204                                                       \\\midrule
\textbf{Proposal Training} & N/A                & 500                                                                                        & 500                                                                                        & 500                                                                   & N/A                                                       \\\midrule
\textbf{Validation}        & 1,497             & 1,497                                                                                     & 1,403                                                                                    & 1,304                                                                   & 1,298                                                       \\\toprule[1.3pt] 
\multicolumn{6}{c}{\textbf{Classes}}                                                                                                                                                                                                                                                                                                                                      \\\toprule[1.3pt]
                           & \textbf{Close-Set} & \textbf{Open-Set}                                                                          & \multicolumn{3}{c}{\textbf{Open-World}}                                                                                                                                                                                    \\\midrule
                           & \textit{Task}     & \textit{Task}                                                                              & \textit{Task 1}                                                                          & \textit{Task 2}                                                    & \textit{Task 3}                                        \\\midrule
\textbf{Known}             & All                & car, truck                                                                            & car, truck                                                                            & car, truck, tram, misc, cyclist                               & car, truck, tram, misc, cyclist, pedestrian, van, person sitting \\\midrule
\textbf{Unknown}           & N/A                & tram, misc, cyclist, pedestrian, van, person sitting & tram, misc, cyclist, pedestrian, van, person sitting & pedestrian, van, person sitting & N/A      \\\bottomrule[1.3pt]
\end{tabular}
\label{tab:kitti_resample}
\end{table*}
The BDD dataset includes over 1,840,000 labeled objects, with a training sample set consisting of 70,000 images, a validation set containing 10,000 images and a test set containing 20,000 images. 
The labeled classes in the dataset are traffic sign, traffic light, pedestrian, car, motor, bus, bike, truck, train, and rider. We exclude the class \textit{train} from our analysis, as it only contains 179 instances, which is significantly lower than the other classes.
\setcounter{table}{2}
\begin{table}[H]
\centering
\caption{Samples and Instances Statistics \\ in BDD100K Dataset}
\begin{tabular}{c|c|c}
\toprule[1.3pt]
\midrule[0.3pt]
\multicolumn{3}{c}{\textbf{Samples}}                             \\\midrule
\textbf{Training}    & \textbf{Validation} & \textbf{Testing}    \\\midrule
70,000               & 10,000              & 20,000              \\\midrule[1.1pt] 
\multicolumn{3}{c}{\textbf{Instances}}                           \\\midrule
\textbf{Classes}     & \textbf{Training}   & \textbf{Validation} \\\midrule
traffic sign              & 175,724              & 34,908              \\\midrule
traffic light           & 159,000             & 26,885              \\\midrule
pedestrian                  & 87,174             & 13,262              \\\midrule
car                 & 443,160              & 102,506              \\\midrule
motor         & 3,002              & 452              \\\midrule
bus                  & 11,672               & 1,597               \\\midrule
bike                & 7,210              & 1,007               \\\midrule
truck                & 28,149              & 4,245               \\\midrule
rider         & 4,517               & 649                       \\\bottomrule[1.1pt]
\end{tabular}
\label{tab:bdd_official}
\end{table}

Table \ref{tab:bdd_official} provides a summary of the instance counts for each class.
The training sample set and validation sample set shown in Table \ref{tab:bdd_official} are used for Close-Set Object Detection since the test sample does not provide ground truth as a reference. On the other hand, for Open-Set and Open-World Object Detection, we train the detector on the nuScenes training sample set and evaluate it on the BDD100K validation sample set. This cross-dataset evaluation helps to assess the detector's generalization performance in a different context.

To resample the dataset, we choose the same classes that exist in both datasets.
For Open-Set, the detector is trained on \textbf{pedestrian} and \textbf{bus} classes in the nuScenes training sample set, while its performance is evaluated on the unknown classes \textbf{truck}, \textbf{bike}, \textbf{car}, and \textbf{motor} in the BDD validation sample set.
For Open-World, \textbf{pedestrian} and \textbf{bus} are introduced in Task 1. In Task 2, \textbf{truck} and \textbf{bike} are newly introduced, and in Task 3, \textbf{car} and \textbf{motor} are newly introduced.
The final resampled dataset is presented in Table \ref{tab:bdd_resample}.

\setcounter{table}{4}
\begin{table*}[h]
\centering
\caption{nuScenes Dataset Resampling}
\begin{tabular}{S|S|L|L|L|L}
\toprule[1.0pt]
\midrule[0.3pt]
\multicolumn{6}{c}{\textbf{Samples}}                                                                                                                                                                                                                                                                                                                              \\ \toprule[1.0pt]
                           & \textbf{Close-Set} & \textbf{Open-Set}                                                                          & \multicolumn{3}{c}{\textbf{Open-World}}                                                                                                                                                                                    \\\midrule
                           & \textit{Task}      & \textit{Task}                                                                              & \textit{Task 1}                                                                            & \textit{Task 2}                                                     & \textit{Task 3}                                         \\\midrule
\textbf{Training}          & 67,000             & 66,500                                                                                     & 16,444                                                                                     & 13,626                                                                   & 12,624                                                       \\\midrule
\textbf{Proposal Training} & N/A                & 500                                                                                        & 500                                                                                        & 500                                                                   & N/A                                                       \\\midrule
\textbf{Validation}        & 16,000             & 16,000                                                                                     & 5,488                                                                                     & 4,548                                                                   & 4,276                                                       \\\toprule[1.3pt] 
\multicolumn{6}{c}{\textbf{Classes}}                                                                                                                                                                                                                                                                                                                                      \\  \toprule[1.3pt]
                           & \textbf{Close-Set} & \textbf{Open-Set}                                                                          & \multicolumn{3}{c}{\textbf{Open-World}}                                                                                                                                                                                    \\\midrule
                           & \textit{Task}     & \textit{Task}                                                                              & \textit{Task 1}                                                                          & \textit{Task 2}                                                    & \textit{Task 3}                                        \\\midrule
\textbf{Known}             & All                & car, pedestrian                                                                            & car, bus                                                                            & car, bus, motor, bike, barrier, traffic cone, road objects                               & car, bus, motor, bike, barrier, traffic cone, road objects, trailer, truck, construction, pedestrian \\\midrule
\textbf{Unknown}           & N/A                & bus, motor, bike, barrier, traffic cone, road objects, trailer, truck, construction & motor, bike, barrier, traffic cone, road objects, trailer, truck, construction, pedestrian & trailer, truck, construction, pedestrian & N/A      \\\bottomrule[1.3pt]
\end{tabular}
\label{tab:nu_resample}
\end{table*}
\setcounter{table}{5}
\begin{table*}[h]
\centering
\caption{BDD Dataset Resampling}
\begin{tabular}{S|S|L|L|L|L}
\toprule[1.3pt]
\midrule[0.3pt]
\multicolumn{6}{c}{\textbf{Samples}}                                                                                                                                                                                                                                                                                                                                      \\\toprule[1.3pt]
                           & \textbf{Close-Set} & \textbf{Open-Set}                                                                          & \multicolumn{3}{c}{\textbf{Open-World}}                                                                                                                                                                                    \\\midrule
                           & \textit{Task}      & \textit{Task}                                                                              & \textit{Task 1}                                                                            & \textit{Task 2}                                                     & \textit{Task 3}                                         \\\midrule
\textbf{Training}          & 70,000             & 66,500 (nuScenes)                                                                                     & 15,688 (nuScenes)                                                               & 15,377 (nuScenes)                                 & 12,688 (nuScenes)                                         \\\midrule
\textbf{Proposal Training} & N/A                & 500 (nuScenes)                                                                                       & 500 (nuScenes)                                                                                        & 500 (nuScenes)                                                                   & N/A                                                       \\\midrule
\textbf{Validation}        & 10,000             & 10,000                                                                                     & 5,219                                                                                     & 3,844                                                                   & 3,178                                                       \\\toprule[1.3pt] 
\multicolumn{6}{c}{\textbf{Classes}}                                                                                                                                                                                                                                                                                                                                      \\  \toprule[1.3pt]
                           & \textbf{Close-Set} & \textbf{Open-Set}                                                                          & \multicolumn{3}{c}{\textbf{Open-World}}                                                                                                                                                                                    \\\midrule
                           & \textit{Task}     & \textit{Task}                                                                              & \textit{Task 1}                                                                          & \textit{Task 2}                                                    & \textit{Task 3}                                        \\\midrule
\textbf{Known}             & All                & pedestrian, bus                                                                            & pedestrian, bus                                                                            & pedestrian, bus, truck, bike           & pedestrian, bus, truck, bike, car, motor \\\midrule
\textbf{Unknown}           & N/A                & truck, bike, car, motor & truck, bike, car, motor & car, motor & N/A      \\\bottomrule[1.3pt]
\end{tabular}
\label{tab:bdd_resample}
\end{table*}

\section{Evaluation Metric}
\subsection{Close-Set and Open-Set Object Detection}
We make 100 detection on each image and adapt the COCO evaluation metrics\cite{lin2014microsoft}, which is based on \textbf{Average Precision} (AP) and \textbf{Average Recall} (AR).
AP measures the accuracy of object detection over all object classes, while AR measures the percentage of correctly detected objects out of the total number of objects in the ground truth over all object classes. 

We compute AP and AR at various IoU thresholds, which is the ratio of the area of overlap between the predicted bounding box and the ground truth bounding box to the area of their union. 
The various IoU thresholds include 0.5, 0.55, 0.6, 0.65, 0.7, 0.75, 0.8, 0.85, 0.9, and 0.95. 
We report mean value of AP and AR across these IoU thresholds, denoted as AP\textsubscript{all} and AR\textsubscript{all}.
In addition to the mean AP across all IoU thresholds, we also report the Average Precision at IoU=0.5 and IoU=0.75, denoted as AP\textsubscript{50} and AP\textsubscript{75}, respectively.

Furthermore, we categorize each object class as small, medium, or large based on its area in pixels as COCO evaluation metrics suggest. Objects with an area less than 32\textsuperscript{2} pixels are considered small, objects with an area between 32\textsuperscript{2} and 96\textsuperscript{2} pixels are considered medium, and objects with an area greater than 96\textsuperscript{2} pixels are considered large.
Based on this classification, we report mean AP and mean AR across all IoU thresholds for small, medium and large objects separately, denoted as AP\textsubscript{s}, AP\textsubscript{m}, AP\textsubscript{l}, AR\textsubscript{s}, AR\textsubscript{m} and AR\textsubscript{l}.
By this way, we can check model's performance on different size objects.

\subsection{Open-World Object Detection}
We report Absolute Open-Set Error (A-OSE) \cite{miller2018dropout} and Wilderness Impact ($WI$)\cite{dhamija2020overlooked} for this problem. A-OSE is the number of unknown objects classified as known objects in error. Lower A-OSE represents better performance. $WI$ evaluates the impact of unknowns objects on the performance of a detector, with a smaller value indicating less impact. The equation to calculate $WI$ is as follows:
\begin{equation}
WI = \frac{P_{K}}{P_{K\cup U}}-1
\end{equation}
Where, $P_{K}$ is the precision at $IoU = 0.5$ evaluated on known classes, $P_{K\cup U}$ is the precision at IoU = 0.5 evaluated on both known and unknown classes.

We also present the AP at IoU=0.5 for known classes to evaluate the model's ability to detect current known classes and previous known classes.
\section{Experiments and Results}
In this section, we conduct experiments on three datasets to demonstrate the effectiveness of the additional saliency map in object detection. These experiments aim to evaluate the proposed method on three different tasks: Close-Set, Open-Set, and Open-World Object Detection.

We use Adam optimizer with $beta1 = 0.5$ and $beta2 = 0.98$ for all experiments. Each batch consists of 16 images, and the learning rate is set to 1e-4. We perform all experiments using NVIDIA A100-80G GPU.

\setcounter{table}{6}
\begin{table*}[hpt]
\centering
\caption{Result of Close-Set Objection Detection}
\begin{tabular}{m{1.2cm}<{\centering}|m{1cm}<{\centering}m{1cm}<{\centering}m{1cm}<{\centering}m{1cm}<{\centering}m{1cm}<{\centering}m{1cm}<{\centering}m{1cm}<{\centering}m{1cm}<{\centering}m{1cm}<{\centering}m{1cm}<{\centering}m{1cm}<{\centering}}

\toprule[1.3pt]
\midrule[0.3pt]
Dataset                   & Method  & AP\textsubscript{all}    & AP\textsubscript{50} & AP\textsubscript{75}  & AP\textsubscript{s} & AP\textsubscript{m} & AP\textsubscript{l} & AR\textsubscript{all} & AR\textsubscript{s} & AR\textsubscript{m} & AR\textsubscript{l}             \\ 
\midrule
\multirow{2}{*}{KITTI}    & $I$    & 0.503          & 0.737          & 0.565          & 0.493          & 0.524          & 0.567          & 0.567          & 0.555          & 0.587          & 0.610 \\ 
                          & $I_S$  & \textbf{0.548} & \textbf{0.748} & \textbf{0.629} & \textbf{0.526} & \textbf{0.573} & \textbf{0.582} & \textbf{0.587} & \textbf{0.560} & \textbf{0.612} & \textbf{0.630}  \\ 
\midrule
\multirow{2}{*}{nuScenes} & $I$    & 0.251          & 0.422          & 0.262          & 0.096          & 0.244          & 0.396          & 0.297          & 0.126          & 0.293          & 0.447           \\ 
                          & $I_S$  & \textbf{0.347} & \textbf{0.536} & \textbf{0.384} & \textbf{0.168} & \textbf{0.346} & \textbf{0.484} & \textbf{0.393} & \textbf{0.203} & \textbf{0.393} & \textbf{0.533}  \\ 
\midrule
\multirow{2}{*}{BDD}      & $I$    & 0.176          & 0.325          & 0.164          & 0.059          & 0.217          & 0.363          & 0.219          & 0.080          & 0.275          & 0.418           \\ 
                          & $I_S$  & \textbf{0.186} & \textbf{0.348} & \textbf{0.174} & \textbf{0.062} & \textbf{0.226} & \textbf{0.401} & \textbf{0.242} & \textbf{0.092} & \textbf{0.298} & \textbf{0.465}  \\ 
\midrule
\multicolumn{12}{l}{$I$ represents the traditional method with original image as input to the detector.}                                                                                                      \\
\multicolumn{12}{l}{$I_S$ represents the proposed method with combination of original image and its saliency map as input to the detector.}  \\
\midrule
\end{tabular}
\end{table*}
\setcounter{table}{7}
\begin{table*}[hpt]
\centering
\caption{Result of Open-Set Objection Detection}
\begin{tabular}{m{1.2cm}<{\centering}|m{1cm}<{\centering}m{1cm}<{\centering}m{1cm}<{\centering}m{1cm}<{\centering}m{1cm}<{\centering}m{1cm}<{\centering}m{1cm}<{\centering}m{1cm}<{\centering}m{1cm}<{\centering}m{1cm}<{\centering}m{1cm}<{\centering}}

\toprule[1.3pt]
\midrule[0.3pt]
Dataset                   & Detector  & AP\textsubscript{all}    & AP\textsubscript{50} & AP\textsubscript{75}  & AP\textsubscript{s} & AP\textsubscript{m} & AP\textsubscript{l} & AR\textsubscript{all} & AR\textsubscript{s} & AR\textsubscript{m} & AR\textsubscript{l}             \\ 
\midrule
\multirow{2}{*}{KITTI}    & OLN     & 0.104          & 0.164          & 0.118          & 0.050           & 0.098          & 0.177          & 0.144          & 0.105          & 0.134          & 0.198 \\ 
                          & SalienDet & \textbf{0.250}  & \textbf{0.484} & \textbf{0.227} & \textbf{0.184} & \textbf{0.234} & \textbf{0.341} & \textbf{0.324} & \textbf{0.236} & \textbf{0.302} & \textbf{0.435}  \\
\midrule
\multirow{2}{*}{nuScenes} & OLN     & 0.021          & 0.039          & 0.017          & 0.008          & 0.018          & 0.036          & 0.038          & 0.006          & 0.036          & 0.058           \\ 
                          & SalienDet & \textbf{0.207} & \textbf{0.429} & \textbf{0.187} & \textbf{0.057} & \textbf{0.202} & \textbf{0.303} & \textbf{0.310}  & \textbf{0.118} & \textbf{0.301} & \textbf{0.425}  \\
\midrule
\multirow{2}{*}{\begin{tabular}[c]{@{}c@{}}nuScenes+\\BDD\end{tabular}}      & OLN     & 0.015          & 0.024          & 0.017          & 0.003          & 0.008          & 0.041          & 0.023          & 0.009          & 0.010           & 0.050            \\
                          & SalienDet & \textbf{0.198} & \textbf{0.417} & \textbf{0.170}  & \textbf{0.038} & \textbf{0.235} & \textbf{0.506} & \textbf{0.279} & \textbf{0.079} & \textbf{0.341} & \textbf{0.601}  \\ 
\midrule
\end{tabular}
\label{Tab:novel_class}
\end{table*}

\subsection{Evaluation for Close-Set Object Detection}
The target of this experiment is to assess whether the use of an additional saliency map could improve the performance of Close-set Object Detection. 
For this experiment, we utilize the Faster R-CNN detector and introduce a modification to the input based on the approach described in Part B of Section II. 
Specifically, we replace the original image tensor with a new image tensor denoted as $I_{s}$. 
This new tensor is created by merging the original images with their corresponding saliency maps, which are generated using the entire image as input.
We conduct a series of comparison using the original image tensor $I$ and the new image tensor $I_{S}$ as inputs on the three datasets. 

These Close-Set detection result is presented in Table VII, which indicates that using the new image tensor $I_{S}$ yields better performance than using the original image tensor on all evaluation metrics, highlighting the advantages of the additional saliency map.
This result demonstrates the feasibility and potential to detect objects by additional highlighted information.   

\subsection{Evaluation for Open-Set Object Detection} 
Detecting new object classes is a crucial aspect of Open-Set Object Detection. 
To investigate whether SalienDet can enhance the detection of such classes, we adopt Kim's experimental methodology \cite{kim2022learning} on the OLN model. 
Our study involves both cross-category and cross-dataset evaluations, following the dataset configurations presented in Tables Table \ref{tab:kitti_resample}, \ref{tab:nu_resample} and \ref{tab:bdd_resample}. Specifically, we conduct cross-category evaluations on KITTI and nuScenes datasets, and cross-dataset evaluations on nuScenes and BDD.

We follow the way described in part C of section III to introduce additional saliency map into detector.

During training, we only utilize annotations from the known classes in the training sample set, and during inference, we evaluate only on the unknown classes in the validation sample set.
 
In addition to presenting the evaluation metrics of SalienDet, we also provide the performance results of the OLN model for comparison purposes.
\begin{figure}[h]
    \centering
    \captionsetup{justification=justified}
    \includegraphics[width=8.5cm]{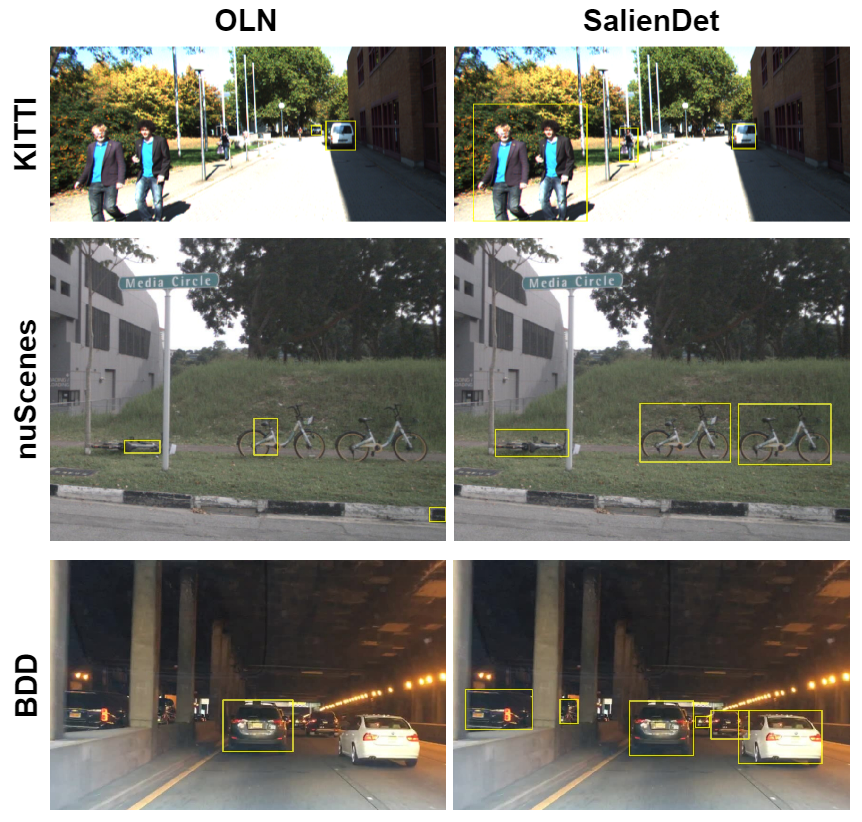}
    \caption{Visualize Comparison between OLN and SalienDet. Yellow box indicates the object proposals from two approaches.}
    \label{fig:novelclass}
\end{figure}

Fig \ref{fig:novelclass} presents the object proposal predictions of the two methods on sample images, and Table \ref{Tab:novel_class} provides the corresponding numerical results. 
As shown in Table \ref{Tab:novel_class}, the OLN model's performance is not satisfactory when the training data contains only two or three object classes. 
However, in Kim's experiment, the OLN model is trained on a set consisting of 20 classes, which provides it with a diverse range of examples to learn the general properties of objects.
In contrast, our method demonstrates improved performance when facing this challenge.

\subsection{Evaluation for Open-World Detection}
In addition to detecting new object classes, incremental learning is a crucial aspect of addressing the Open-World Object Detection problem. To tackle this challenge, we adopt the proposed dataset modification approach described in Part D of Section III. To provide a comprehensive evaluation of our approach, we conduct both within-dataset and cross-dataset evaluations based on the dataset configurations presented in Tables \ref{tab:kitti_resample}, \ref{tab:nu_resample}, and \ref{tab:bdd_resample}.
\setcounter{table}{8}
\begin{table*}[hpt]
\centering
\caption{Result of Open-World objection Detection}
\begin{tabular}{cc|ccc|ccccc|ccc} 
\toprule[1.3pt]
\midrule[0.3pt]
            & \multicolumn{1}{c}{}      & \multicolumn{3}{c|}{Task1}                                                                            & \multicolumn{5}{c|}{Task2}                                                                                                                                                       & \multicolumn{3}{c}{Task3}                                                                                                          \\\midrule 
\toprule[1.3pt]
\multirow{2}{*}{Dataset}                                                                 & \multirow{2}{*}{Detector} & \multirow{2}{*}{WI $\left( \downarrow{} \right)$} & \multirow{2}{*}{A-OSE $\left( \downarrow{} \right)$} & mAP                                                    & \multirow{2}{*}{WI $\left( \downarrow{} \right)$} & \multirow{2}{*}{A-OSE $\left( \downarrow{} \right)$}  & \multicolumn{3}{c|}{mAP}                                                                                                           & \multicolumn{3}{c}{mAP}                                                                                                            \\ \cline{5-5}\cline{8-13}
     &                   &           &             & \begin{tabular}[c]{@{}c@{}}current\\known \end{tabular} &                     &                        & \begin{tabular}[c]{@{}c@{}}previous\\known\end{tabular} & \begin{tabular}[c]{@{}c@{}}current\\known\end{tabular} & both           & \begin{tabular}[c]{@{}c@{}}previous\\known\end{tabular} & \begin{tabular}[c]{@{}c@{}}current\\known\end{tabular} & both            \\ \midrule

\multirow{2}{*}{KITTI}                                                  & Baseline                  & 0.332               & 1030                   & 0.781                                                  & 0.106               & 1017                   & 0.704                                                   & 0.26                                                   & 0.437          & 0.373                                                   & 0.356                                                  & 0.368           \\
                                                                        & SalienDet                 & \textbf{0.150}      & \textbf{907}           & \textbf{0.866}                                         & \textbf{0.033}      & \textbf{926}           & \textbf{0.736}                                          & \textbf{0.341}                                         & \textbf{0.499} & \textbf{0.443}                                          & \textbf{0.484}                                         & \textbf{0.455}  \\ 
\midrule
\multirow{2}{*}{nuScenes}                                               & Baseline                  & 0.351               & 4240                   & 0.370                                                  & 0.113               & 3537                   & 0.292                                                   & 0.413                                                  & 0.372          & 0.286                                                   & 0.154                                                  & 0.238           \\
                                                                        & SalienDet                 & \textbf{0.022}      & \textbf{3461}          & \textbf{0.475}                                         & \textbf{0.005}      & \textbf{2816}          & \textbf{0.352}                                          & \textbf{0.464}                                         & \textbf{0.430} & \textbf{0.310}                                          & \textbf{0.242}                                         & \textbf{0.285}  \\ 
\midrule
\multirow{2}{*}{\begin{tabular}[c]{@{}c@{}}nuScenes+\\BDD\end{tabular}} & Baseline                  & 0.316               & 7885                   & 0.281                                                  & 0.200               & 22661                  & 0.100                                                   & 0.102                                                  & 0.103          & 0.112                                                   & 0.076                                                  & 0.099           \\
                                                                        & SalienDet                 & \textbf{0.079}      & \textbf{2001}          & \textbf{0.285}                                         & \textbf{0.065}      & \textbf{5419}          & \textbf{0.112}                                          & \textbf{0.164}                                         & \textbf{0.133} & \textbf{0.144}                                          & \textbf{0.116}                                         & \textbf{0.133}  \\
\midrule
\end{tabular}
\label{tab:open_world_result}
\end{table*}

\begin{figure}[h]
    \centering
    \captionsetup{justification=justified}
    \includegraphics[width=8.5cm]{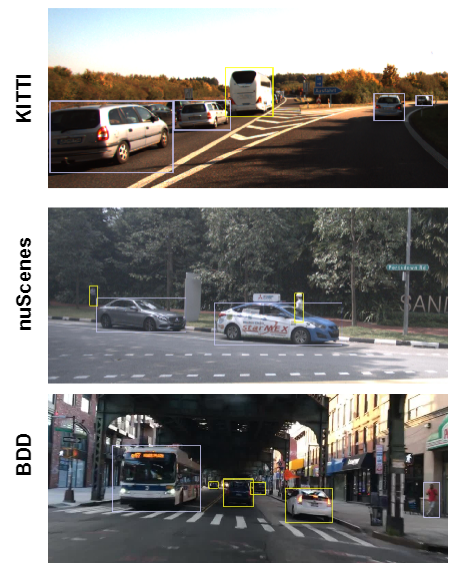}
    \caption{Predictions after being trained with modified dataset on Task 1. The unknown objects in yellow and known objects in pink are classified correctly. }
    \label{fig:owd_result}
\end{figure}
Furthermore, we used the same dataset configuration to train a Faster R-CNN detector for comparison. To ensure a fair comparison, we also allowed the Faster R-CNN detector to fine-tune using the proposal training sample set that SalienDet uses as prior knowledge. This means that the detector also has some prior knowledge about the objects in the dataset, but its ability to detect unknown objects is still limited.

In Table \ref{tab:open_world_result}, we present a comparison between our proposed approach and Faster R-CNN with fine-tuning. Fig \ref{fig:owd_result} shows the visualization result of our proposed approach.

Our proposed approach outperforms the comparison model in terms of lower $WI$ and A-OSE. 
By utilizing a modified training sample set, our detector has already learned the properties of potential unknown objects in a previous task. Therefore, when some of these potential objects are introduced as new classes in the training set of a new task, our detector can continue learning from these newly labeled objects and extract better features from images. As a result, it performs better in detecting these newly introduced objects compared to the comparison model.
Since the detector has been trained on a dataset that includes both known and unknown objects, it has already learned how to extract features that differentiate between them. 
Even if the detector is only presented with newly introduced classes in a later task, it still has a stronger ability to extract features that show the difference between the previously known classes and newly introduced classes. 
This results in the detector having a better ability to recall previously learned objects.

\subsection{Ablation Study}
\setcounter{table}{9}
\begin{table*}[h]
\centering
\caption{Comparison of Results with and without Additional Saliency Map}
\begin{tabular}{m{1.5cm}<{\centering}m{1cm}<{\centering}m{1cm}<{\centering}m{1cm}<{\centering}m{1cm}<{\centering}m{1cm}<{\centering}m{1cm}<{\centering}m{1cm}<{\centering}m{1cm}<{\centering}m{1cm}<{\centering}m{1cm}<{\centering}}

\toprule[1.3pt]
\midrule[0.3pt]
   & AP\textsubscript{all}    & AP\textsubscript{50} & AP\textsubscript{75}  & AP\textsubscript{s} & AP\textsubscript{m} & AP\textsubscript{l} & AR\textsubscript{all} & AR\textsubscript{s} & AR\textsubscript{m} & AR\textsubscript{l}             \\ 
\midrule
\multicolumn{1}{c|}{Without} & 0.034          & 0.069          & 0.027          & 0.009          & 0.026          & 0.061          & 0.049          & 0.012          & 0.035          & 0.086           \\
\midrule
\multicolumn{1}{c|}{With}    & \textbf{0.207} & \textbf{0.429} & \textbf{0.187} & \textbf{0.057} & \textbf{0.201} & \textbf{0.303} & \textbf{0.310} & \textbf{0.118} & \textbf{0.301} & \textbf{0.425}  \\
\midrule
\end{tabular}
\label{Tab:ablation1}
\end{table*}
\setcounter{table}{10}
\begin{table*}[h]
\centering
\caption{Impact of Different Backbone Architectures on Detection Performance in SalienDet}
\begin{tabular}{m{1.5cm}<{\centering}|m{1cm}<{\centering}m{1cm}<{\centering}m{1cm}<{\centering}m{1cm}<{\centering}m{1cm}<{\centering}m{1cm}<{\centering}m{1cm}<{\centering}m{1cm}<{\centering}m{1cm}<{\centering}m{1cm}<{\centering}}

\toprule[1.3pt]
\midrule[0.3pt]
 Backbone  & AP\textsubscript{all}    & AP\textsubscript{50} & AP\textsubscript{75}  & AP\textsubscript{s} & AP\textsubscript{m} & AP\textsubscript{l} & AR\textsubscript{all} & AR\textsubscript{s} & AR\textsubscript{m} & AR\textsubscript{l}             \\ 
\midrule
Resnet-50       & 0.207 & 0.429 & \textbf{0.187} & 0.057 & 0.201 & 0.303 & 0.310  & \textbf{0.118} & 0.301 & 0.425  \\
Resnet-101      & \textbf{0.212} & \textbf{0.448} & 0.186 & \textbf{0.058} & \textbf{0.202} & \textbf{0.313} & \textbf{0.324}  & 0.114 & 0.317 & \textbf{0.448}  \\
Darknet-53      & 0.206 & 0.430 & 0.184 & 0.054 & 0.206 & 0.307 & 0.320  & 0.117 & \textbf{0.319} & 0.434  \\
Swin-T & 0.190 & 0.397 & 0.167 & 0.048 & 0.177 & 0.287 & 0.297  & 0.110 & 0.292 & 0.406  \\
\midrule
\end{tabular}
\label{Tab:ablationa}
\end{table*}
We conduct an ablation study on the SalienDet. 
First, we examine the impact of the additional saliency maps on detecting new object classes by training the detector without incorporating saliency maps. 
Then, we evaluate the effect of different backbones on performance while incorporating the additional saliency maps into the detector \cite{ghorai2022state}. 
We use Resnet-101 \cite{he2016deep}, Darknet-53 \cite{redmon2018yolov3}, and Swin Transformer (Swin-T) \cite{liu2021swin} as the alternative backbones, in addition to the original Resnet-50. 
The experiment is conducted on the nuScenes dataset configuration shown in \ref{tab:nu_resample} , and the other experimental settings were consistent with Part B of Section VI. 
Table \ref{Tab:ablation1} and \ref{Tab:ablationa} present the results of this study.

As shown in Table \ref{Tab:ablation1}, the object detector performs poorly in detecting novel class objects without the use of an additional saliency map. 
However, the incorporation of the saliency map results in significant improvement in detecting novel objects.
Specifically, the AP\textsubscript{50} metric increases from 0.069 to 0.429, indicating a substantial enhancement in the detector's performance.

Table \ref{Tab:ablationa} shows that the ResNet-101 backbone achieves the highest performance in terms of various metrics, including AP\textsubscript{all}, AP\textsubscript{50}, AP\textsubscript{s}, AP\textsubscript{m}, AP\textsubscript{l}, and AR\textsubscript{l}. 
This suggests that SalienDet can perform better with a CNN-based backbone architecture that has more layers. 
However, it is interesting to note that the Swin Transformer, which is a transformer-based architecture and has shown state-of-the-art performance in Close-Set Object Detection \cite{liu2021swin}, performs the worst compared to other backbones. 
While its attention mechanism can selectively attend to informative regions related to unknown objects, it may not be as effective as other backbones in detecting unknown classes due to a tendency to focus more on informative regions related to known classes.

\section{CONCLUSIONS and FUTURE WORK}

This paper investigates the efficacy of incorporating additional saliency map information in OD, and evaluates its performance on three autonomous vehicle datasets. For Close-Set problem, we find that incorporating the additional saliency map with the original image leads to an improvement in detection performance through investigation;
for Open-Set problem, we introduce a novel solution called SalienDet that utilizes an additional saliency map to detect unknown objects, which outperforms the comparison model on three datasets;
for Open-World problem, we propose a dataset modification approach, which utilizes the SalienDet. With the modified dataset, a regular OD application performs well in both detecting unknown objects and incremental learning tasks;
finally, we conduct the ablation study, demonstrating the effectiveness of the additional saliency map on detecting unknown objects.

However, our proposed SalienDet still has some limitations that require future research to address. One major challenge is that SalienDet relies on prior knowledge to generate proposals for the additional saliency map in this study, which may not always be available in practice. Another issue is that although the Swin Transformer is a strong backbone and performs well on Close-Set problem, it does not perform as well in detecting novel class objects, limiting its effectiveness in our study. It is unfortunate that we could not utilize more advanced methods in this study. We hope that future work can overcome these challenges and further improve the performance of SalienDet, making it more comparable to human-like detection capabilities.

\bibliographystyle{unsrt}
\bibliography{ref}

\addtolength{\textheight}{-12cm}
\begin{wrapfigure}{l}{25mm} 

\includegraphics[width=1in,height=1.5in,clip,keepaspectratio]{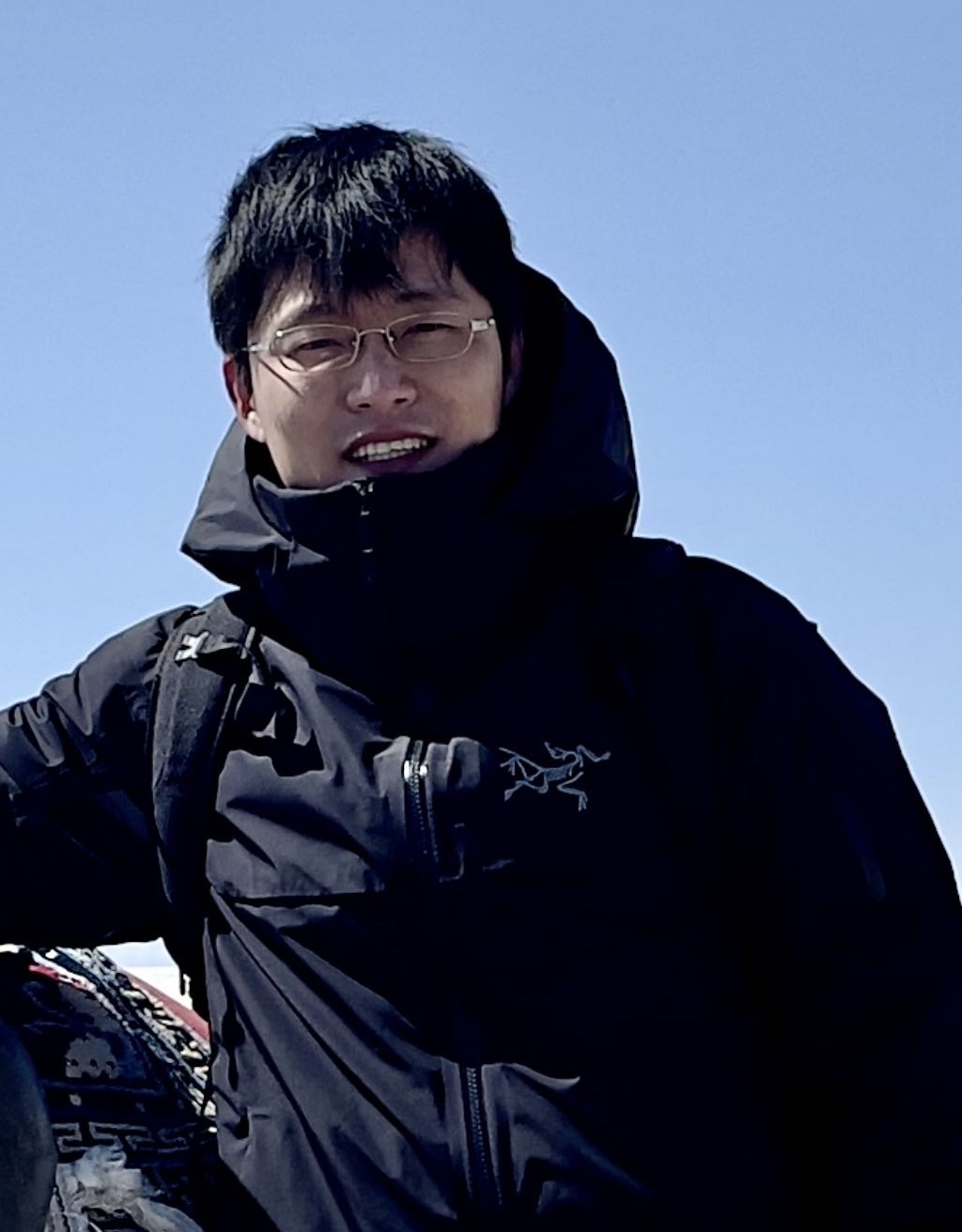}

\end{wrapfigure}\par

\textbf{Ning Ding} received his B.S. degree and M.S. degree, both in Vehicle Engineering (Automobile) from Tongji University, China. He is currently pursuing a Ph.D. degree in mechanical engineering at Virginia Tech Autonomous System and Intelligent Machine (ASIM) Lab. His research interests include autonomous vehicle, machine learning, human machine interface.\par
\hfill

\begin{wrapfigure}{l}{25mm}
\includegraphics[width=1in,height=1.5in,clip,keepaspectratio]{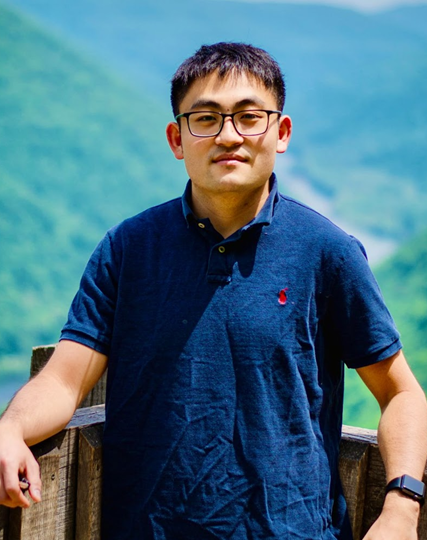}
\end{wrapfigure}\par
\textbf{Ce Zhang} earned his B.Sc degree in mechanical engineering from Virginia Tech in 2018. He then earned his Ph.D. degree in mechanical engineering from Virginia Tech in 2023. His research interests include autonomous vehicle perception, computer vision, neural networks, and electroencephalography (EEG). He is a member of IEEE.\par
\hfill

\begin{wrapfigure}{l}{25mm}
\includegraphics[width=1in,height=1.5in,clip,keepaspectratio]{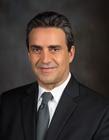}
\end{wrapfigure}\par
\textbf {Azim Eskandarian} has been a Professor and Head of the Mechanical Engineering Department at Virginia Tech (VT) since August 2015. He became the Nicholas and Rebecca Des Champs chaired Professor in April 2018. He established the Autonomous Systems and Intelligent Machines laboratory at VT to conduct research in intelligent and autonomous vehicles, and mobile robotics. Prior to that, he was a Professor of Engineering and Applied Science at The George Washington University (GWU) and the founding Director of the Center for Intelligent Systems Research (1996-2015), the director of the Transportation Safety and Security University Area of Excellence (2002-2015), and the co-founder of the National Crash Analysis Center (1992) and its Director (1998-2002,  5/2013-7/2015). Earlier, he was an Assistant Professor at Pennsylvania State University, York, PA (1989-92) and worked as an engineer/project manager in industry (1983-89). He is the Editor-in-Chief of the IEEE Transactions on Intelligent Transportation Systems. He was awarded the IEEE ITS Society’s Outstanding Researcher Award in 2017 and the GWUs School of Engineering Outstanding Researcher Award in 2013. Dr. Eskandarian is a fellow of ASME, senior member of IEEE, and member of SAE professional societies. He received his BS, MS, and DSC degrees in Mechanical engineering from GWU, Virginia Tech, and GWU, respectively.
\par

\end{document}